# Judicial Intelligent Assistant System: Extracting Events from Divorce Cases to Detect Disputes for the Judge

Yuan Zhang[a], Chuanyi Li[a*], Yu Sheng[a], Jidong Ge[a,b], Bin Luo[a]

[a] *State Key Laboratory for Novel Software Technology, Software Institute, Nanjing University, China*

[b] *State Key Laboratory of Networking and Switching Technology, Beijing University of Posts and Telecommunications, China*

*Corresponding Author Email: lcy@nju.edu.cn

*Abstract*—In formal procedure of civil cases, the textual materials provided by different parties describe the development process of the cases. It is a difficult but necessary task to extract the key information for the cases from these textual materials and to clarify the dispute focus of related parties. Currently, officers read the materials manually and use methods, such as keyword searching and regular matching, to get the target information. These approaches are time-consuming and heavily depending on prior knowledge and carefulness of the officers. To assist the officers to enhance working efficiency and accuracy, we propose an approach to detect disputes from divorce cases based on a two-round-labeling event extracting technique in this paper. We implement the Judicial Intelligent Assistant (JIA) system according to the proposed approach to 1) automatically extract focus events from divorce case materials, 2) align events by identifying co-reference among them, and 3) detect conflicts among events brought by the plaintiff and the defendant. With the JIA system, it is convenient for judges to determine the disputed issues. Experimental results demonstrate that the proposed approach and system can obtain the focus of cases and detect conflicts more effectively and efficiently comparing with existing method.[1]

***Keywords***: Expert Systems, Information Retrieval, Event Extraction, Conflict Detection, Judicial Procedure.

## 1. Introduction

In a Chinese civil litigation case, there are usually some important textual litigation materials which are provide by the parties or surveyed by court officers. Further judicial works are based on these materials. For example, the judges could learn the development process of a case from them to conduct conciliation or judgment referring to laws. For each case, the judges should sort out key facts of it, as well as the dispute issues between different parties. However, this mainly rely on reading these materials manually, and use methods, such as keyword searching or regular matching (Merrouni, Frikh, & Ouhbi, 2019), to get core information at present. But there are many drawbacks. First, it is labor-intensive and time-consuming. Second, judges should focus on different key information while dealing with various cases, which caused by different types of reasons. This require they ought to have a wealth of priori knowledge of different types of cases. At last, manually searching or matching depend heavily on the carefulness and experience of the judges, which are of great uncertainty and instability. If there is an approach to automatically acquire key information from the materials of cases accurately and to accelerate identify disputed issues, both labor and time costs would be reduced. It would also benefit further works if we store both the structured information and original materials in the data base (Hao, Wang, & Ng, 1996). Besides, to some extent, the understanding of cases and standards of the trial would be unified.

Considering that the case materials are described in natural language, it's difficult to derive key information of cases automatically only by keyword retrieving or regular matching. So, in this paper, we propose a novel two-round-labeling event extracting method for retrieving key facts described by different parties. Based on this method, we design a dispute detecting approach for assisting judges handling divorce cases more effectively and efficiently. We also implement the corresponding expert system, named Judicial Intelligent Assistant (JIA) system. According to the uploaded litigation materials containing the statements of the plaintiff and the defendant of the case, the system would extract key events of predefined types about the couple's marriage and identify whether there are conflicts between the events of a same type providing by different parties. The judges would quickly understand the case through the list of events and clarify the dispute issues between the parties through the conflicts.

---

[1] An earlier version (six-pages double-column) of this paper was presented at the 2019 ACM International Joint Conference on Pervasive and Ubiquitous Computing and the 2019 ACM International Symposium on Wearable Computers, UbiComp/ISWC 2019 Adjunct, i.e., Li, Sheng, Ge & Luo (2019). This paper is essentially different from the previous version in the following aspects: (1) **Goal**: the goal of this paper is to detect disputes in civil cases, while the previous paper is to extract events from case descriptions. For achieving the new goal, this paper proposes novel methods for events alignment and disputes detection, and evaluates them with comprehensive experiments. (2) **Procedure of defining Event Types**: the process of defining event types is redesigned in this paper and they are described more detailed than in the previous version. (3) **Events definition**: attributes of the 13 events proposed in the previous paper are reorganized in this paper. (4) **Data annotation**: the entire dataset is re-annotated according to the new event definitions. (5) **Event extracting approach**: the list of keywords for locating events is expended and the model for automatically extracting events is changed. (6) **Experiments**: event extraction experiments are reconducted and new results are reported.

The process for implementing the JIA system is mainly divided into three parts. The first part is to define the types and contents of events that should be extracted. Although there are plenty of definitions of event types in existing event extraction tasks, such as Message Understanding Conferences (MUC), Automated Content Extraction (ACE) and Text Analysis Conferences (TAC), there is not a group of events that can exactly cover or match the focus in divorce cases. So firstly, we propose a detailed portable mechanism to discover and define concrete event types from the focus of judicial cases based on the procedure defined in Li, Sheng, Ge & Luo (2019). Then event types of divorce cases are defined with this mechanism because that the JIA system is to processing divorce cases. The second part is to design a proper event extraction approach according to the defined events. Our proposed specific event extracting approach is characterized by a two-round-labeling strategy which is used for handling the problem of sentences containing multiple events sharing same arguments or trigger words. Existing approaches, such as (Chen et al., 2017; Chen et al., 2015; Liu, Luo, & Huang, 2018), can't solve the problem very well. The last part is to align co-reference events extracted from statements of different parties and then detect conflicts. This helps reduce the complexity of detecting the disputed issues between parties.

For a comprehensive evaluation of the JIA system, we investigate the accuracy of the proposed event extracting approach and the effectiveness/efficiency of the implemented system separately. The experimental results based on manually annotated data shows that the proposed two-round-labeling event extracting strategy outperforms the existing methods in civil divorce cases. The experiments of comparing results derived by processing the case materials manually and with JIA system show that our system improves the efficiency of getting the focus of cases and disputes between parties. To conclude, we made the following contributions in this paper:

(1) Propose a portable mechanism for discovering and defining event types for judicial cases. Specifically, we define 13 types of events for divorce cases, as well as a list of trigger words.

(2) Design a two-round-labeling approach to extract events from case materials and implement the corresponding Judicial Intelligent Assistant system. To the best of our knowledge, this is the first work adopting event extraction technique to detect dispute issues in the judicial field.

(3) Manually annotate textual materials of 3100 divorce cases according to event definition to construct a public and reusable research data set[2].

The remainder of this paper is structured as follows. Section 2 provides the basic knowledge about Chinese judicial litigation and the formal foundations about event extraction. Section 3 illustrates the concrete approach for getting focus events and helping determine disputes from divorce cases materials, including defining event types, extracting events and detecting conflicts between parties. Section 4 evaluates the effectiveness of the proposed two-round-labeling approach and the JIA system[3]. Section 5 introduces some state-of-the-art methods of event extractions as related works. Section 6 concludes the paper and describes the future work.

**2. Background**

In this section, we introduce the formal processing procedure of civil cases in the People's Republic of China, basic knowledge of event extraction, and the challenges while applying event extraction in handling textual material of divorce cases.

*2.1. Formal Procedure of first round trial*

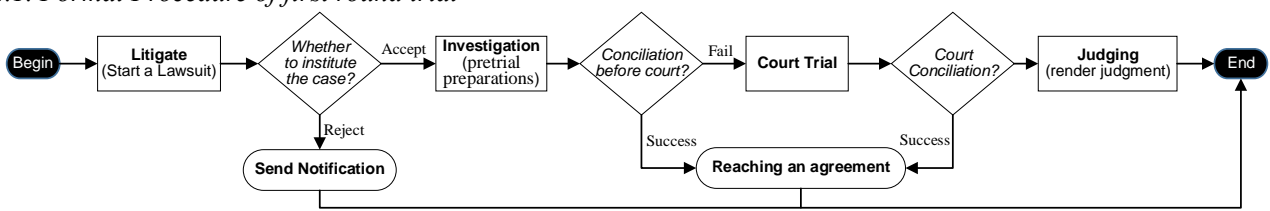

**Figure 1**. Formal Procedure at First Instance

The procedure of first round trial for civil cases, i.e., Formal Procedure at First Instance, refers to the procedure normally used by the court in Peoples' Republic of China to hear and adjudicate civil cases during the first-round trial. It has the following key steps (*The Civil Procedure Law of the People's Republic of China, 2017 Revision*), as is shown in Figure 1:

● **Litigate** (start a lawsuit). Any citizen, legal representative or organization that has a direct relation in the case can bring a lawsuit and submit a statement of complaint to the people's court. A people's court place the case on the docket if it meets the standards, otherwise reject it.

---

[2] http://lichuanyi.info/files/resources_of_eedd/Labeled_Dataset_for_event_extracting_of_divorce_cases.zip
[3] Source code can be found here: http://lichuanyi.info/files/resources_of_eedd/znwszz_v2_add_dispute_detect.zip

- **Investigation** (pretrial preparations). After docketing a case, the people's court shall send a copy of the statement of complaint to the defendant. The defendant shall file a defense from the receipt of the copy of the statement of complaint. The judicial officers must carefully examine and verify the case materials and carry out investigations to collect the necessary evidence.
- **Court trial**. The people's court shall, whenever necessary, go on circuit to hold trials on the spot. The court clerk shall make a written record of the entire court proceedings.
- **Judging** (render judgment). In trying civil cases, the people's courts shall conduct conciliation for the parties on a voluntary and lawful basis; if conciliation fails, judgments shall be rendered without delay.

Our proposed approach mainly serves in the ***Investigation*** and ***Court trial*** stage. During the investigation, plaintiff's complaint and defendant's answers to complaint are collected. Then officers should obtain key information of the case according to their knowledge of law to determine whether to docket the case. The proposed approach and corresponding system would automatically extract key events described by different parties and detect their conflicts. This helps officers to understand the case and detect the focus of disputes quickly, which plays an important role in further judging the case.

Here is a simple example of the complaint statement made by the plaintiff and the corresponding answer made by the defendant:

- **Plaintiff**: "… *Due to our lack of understanding before marriage, the relationship between us after marriage is average. The defendant had a heavy drinking and smoking **habit**, and had an arrogant personality, a romantic nature, and a irritable temper. He often **quarreled** with me because of life chores. Since 2007, the defendant has **lived together** with another woman. For many times, after coming back home in the night, he **provoked** and **scolded** me for no reason, and then **kicking me out** of the house. …*"

- **Defendant**: "… *I am responsible for my family. I am honest and have **no bad habits**. Regarding the plaintiff's unreasonable request and making trouble, I try to be as **tolerant** as possible. The plaintiff had nothing out of nothing, suspected me for no reason, and artificially created family conflicts. I have **no improper relationship** with other women, nor have I had children with other women. …*"

In the investigating period, upon the judge deriving the statements, they should refer to laws to extract critical information from statements of both sides, and then check if there are conflicts between them. The conflicts would give instructions to the judge in collecting evidences. However, the statements provided by both sides are often too lengthy to read and aligning key information from both sides is a heavy task. But the proposed JIA system would help the judges retrieve all critical events described in the statements according to the predefined event types. Then automatically aligning them for detecting conflicts.

For this example, the goal of the JIA system is to derive events related to words of ***BOLD-ITALIC-UNDERLINE*** in the statements, align them, and check if there are conflicts between them. The judge could view both the initial statements and the processed results in the system.

*2.2. Event and Event Extraction*

Event extraction (EE) is a specific subtask of Information Extraction (IE) which is the task of automatically extracting structured information from unstructured and/or semi-structured machine-readable documents. Automated Content Extraction (ACE) defined that an event is a specific occurrence involving participants (Grishman, Westbrook, & Meyers, 2005). And an event usually consists of a trigger that identifies the occurrence of the event and a series of arguments (including participants and attributes) that server different roles in the event. ACE 2005[4] defines 8 types of events which include 33 subtypes. Event Extraction is the process of determining whether an event is contained in a natural text statement and finding out the event trigger and event arguments, both of which is usually a word chunk in Chinese. And many studies divide event extraction into four subtasks: trigger identification, trigger type determination, argument identification and role (argument) determination (Li & Zhou, 2015).

Event Extraction approaches can be mainly classified into knowledge-driven methods, data-driven methods and hybrid methods (Fei, Ren, & Ji, 2020). Knowledge-driven methods is often based on patterns that express the expert knowledge. It's widely used in the industry due to its good interpretability and high accuracy in specific fields (Valenzuela-Escárcega, Hahn-Powell, Surdeanu, & Hicks, 2015). However, it's often has a low recall and needs to re-establish rules when applied to new fields. It's labor-extensive and time-consuming. Data-driven methods can be divided into two categories: traditional feature-based approaches and deep neural network-based approaches. Feature-based approaches usually require effort to develop rich sets of features which are usually generated by some

---

[4] http://www.ldc.upenn.edu /collaborations/past-projects/ace

Natural Language Processing (NLP) toolkits. However, using NLP toolkits may lead to severe error propagation and limit the application of the models to languages for which such NLP tools are available. The deep neural network can automatically obtain the potential correlation between the inputs, it has a strong generalization ability and self-learning ability. And thanks to the rapid increase in computing power, it is possible to apply deep neural networks to NLP and the last couple of years witness the success of the deep neural network models for event extraction, such as Chen et al. (2015), Feng, Qin, & Liu (2016), Huang et al. (2018), and Wiedmann (2017). But there are two drawbacks of the data-driven methods to event extraction. One is that they do not deal with meaning explicitly, another is that a large amount of data is required in order to get statistically significant results. As both approaches have their disadvantages, some research use hybrid event-extraction methods which combine knowledge-driven methods and data-driven methods to yield the best results (Lei, Zhang, & Liu, 2005).

In general, there are many methods of event extraction, each of which has its advantages and disadvantages. In practical use, it is necessary to balance various requirements according to specific domains, choose appropriate methods and solve some domain-specific problems. In our approach, we adopt the hybrid event extraction method for detecting critical events in divorce cases for the judges. The event types and trigger words are defined according to domain knowledge. The deep neural networks are used for extracting semantic features of sentences, while some traditional hand-crafted features are also adopted.

*2.3. Challenges in Extracting Events from Divorce Cases*

Existing event extraction techniques have been proved to be effective in some respective research fields. But when applying event extraction techniques to the judicial field, there are some challenges as follows:

● **Mismatching of Event Types**. ACE 2005 defines only 33 event types, which do not exactly match the type of event to be concerned about when hearing a case. For example, when hearing a divorce dispute case, it's necessary to consider whether the plaintiff has ever filed a suit again within six months in the absence of any new developments or new reasons, which shall not be entertained. ACE defines the type of Justice-Sue, but it's not limited to divorce proceeding. There are also some judicial event types that do not fall into these 33 event types.

● **Lacking of Proper Data Sets**. Many extraction technologies are experimentally verified on existing publicly available English data sets. As far as we know, this is the first time to attempt to apply event extraction technologies to Chinese judicial field. Therefore, there is no standard experimental data. And some cases, such that (Wiedmann, 2017; Yang et al., 2018; Zeng et al., 2018) make use of Distance Supervision (DS) to automatically generate many training data. But for the target events in the judicial field, there is no existing structured database that can support the DS technology.

● **Sharing Trigger Words and Attributes in Multiple Events**. In case materials, there are many sentences containing multiple events that share arguments or trigger words. As shown in Figure 2(a) and (b) (where Trigger chunks and arguments are in **bold**), there are two events ("met" and "married") in E1, where the two events share Time and Participant arguments. In E2, although the trigger chunk "gave birth to" only appears once, but we can judge that there are two "Be-Born" events according to the following text "the eldest son Ping A" and "the second son Ping B". Most existing event technologies focus on extracting a single event from a single sentence. Even if some researchers consider this phenomenon, they either do not solve the problem or their solutions are not good in our data, for example Liu et al. (2018) and Zeng et al. (2018), which we will be analyzed in detail in Section 4.

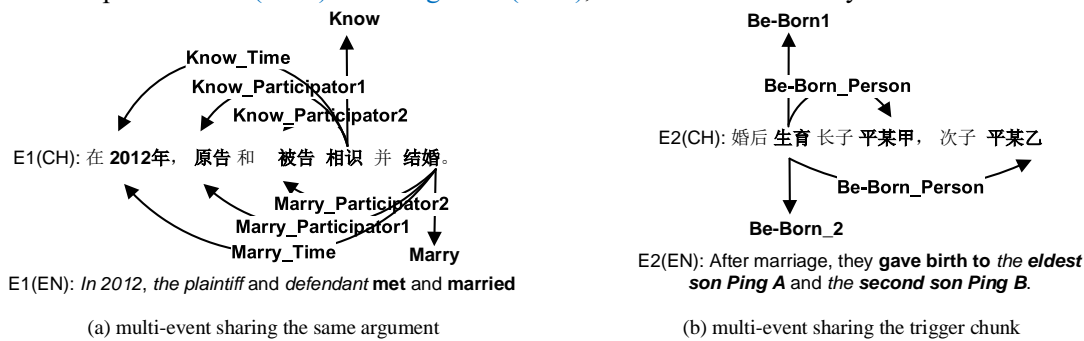

(a) multi-event sharing the same argument    (b) multi-event sharing the trigger chunk

**Figure 2**. Examples of multi-event sharing the arguments or trigger chunk in a single sentence.

To overcome these challenges, we propose an approach for defining the core events, extracting these events from textual materials and further checking conflicts. The implementation of this approach is described in detail in the next section.

## 3. Approach

In this section, we introduce our approach for assisting the trial of divorce cases in detail. Subsection 3.1 gives an overview of the approach and each subsection from 3.2 to 3.6 introduces one part of the approach.

*3.1. Overview*

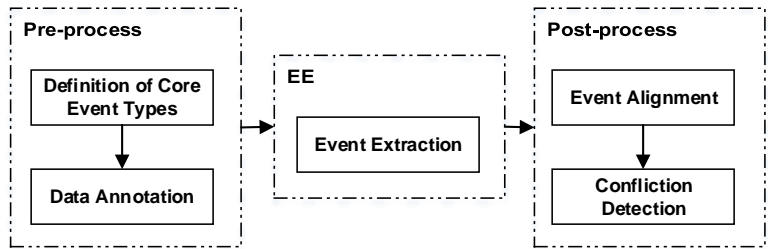

**Figure 3**. Overview of the Approach

Figure 3 show the framework of our approach. The approach consists of five components:

● Definition of core events. We propose a mechanism for defining core events which contain key information about a specific cause of action. And we defined the core events of divorce case.

● Data annotation. The existing technology, DS, for automatically generating training data cannot be implemented because of the lack of existing database with knowledge of the predefined events. So, we manually annotated a bunch of data, the litigation materials for 3100 cases.

● Event extraction. Extract the predefined core events from case materials and solve the problem of multiple event sharing the same trigger or arguments chunk in a single sentence.

● Event alignment. Every type of event may have more than one instance, and each event instance may have more than one mentions., so we need to identify co-reference event mentions that describe the same event instance.

● Event conflictiction detection. Detect whether the co-reference event mentions are inconsistency. If there is, the facts presented by the plaintiff and defendant are in conflict, which may be the disputed issues between the two parties.

*3.2. Define Event Types*

**3.2.1. Collect Focus Set**

In the process of trial, the focus of cases with different causes is different, and the core events are also different. But the focus of any types of case can be found in existing laws. For example, the article 111 of *The Civil Procedure Law of the People's Republic of China* reads:

*"… in a divorce case in which a judgment has been made disallowing the divorce, or in which both parties have become reconciled after conciliation, or in a case concerning adoptive relationship in which a judgment has been made or conciliation has been successfully conducted to maintain the adoptive relation-ship, if the plaintiff files a suit again within six months in the absence of any new developments or new reasons, it shall not be entertained."*

There are three focus could be generated:

● Whether either party has filed for divorce before, and if yes,

  ● whether there is any new developments or new reasons,

  ● whether it is more than six months ago.

For another example，the article 32 of *Marriage Law of the People's Republic of China* reads:

*"In dealing with a divorce case, the people's court should carry out mediation between the parties. Divorce shall be granted if mediation fails because mutual affection no long exists. Divorce shall be granted if mediation fails under any of the following circumstances: (1) bigamy or, cohabitation of a married person with any third party. (2) domestic violence or, maltreatment and desertion of one family member by another; (3) bad habits of gamble or drug addiction which remain incorrigible despite repeated admonition; (4) separation caused by incompatibility, which lasts two full years; and (5) any other circumstances causing alienation of mutual affection."*

It indicates that an important factor in whether a judge decides to allow a divorce is whether there is still affection between the two parties, and lists a number of circumstances causing alienation of mutual affection. So, we get following focus:

● Whether the marriage parties have bigamy or cohabitation.

● Whether there is domestic violence.

● Whether the marriage parties have drug abuse, gambling and other vices.

● Whether or not to be separated. Whether or not to be separated for more than 2 years.

● Whether the mutual affection has been alienated.

Therefore, all the focus can be obtained through the laws and some documents interpreting the laws, such as *Interpretations of the Supreme People's Court about Several Issues Concerning the Application of the Marriage Law of the People's Republic of China (I), (II), (III)*.

However, the focuses derived from the statutes may be implied by the statements of both parties. For example, considering the focus "whether there is any new developments or new reasons", the judge could never determine the focus on the basis of the materials of current case alone. So, after the focuses are generated, they need to be adjusted, i.e., removing the focus that cannot be obtained from the litigation materials. After that, we show the focus collection to the judges who frequently hear divorce cases and ask them to determine whether the collection of focus have included the key information which are normally be focused on in their daily trial. According to their feedback, we filtering or editing the set of focus again.

### 3.2.2. Convert Focus to Event Types

After obtaining the final set of focus, we need to convert the focus into extractable event structures in two ways:

● Direct conversion. For example, the focus-whether there is domestic violence. It can be directly converted into a **Domestic-Violence** event (including three arguments *Perpetrator*, *Victim*, and *Violence Time*).

● Indirect conversion. For example, the focus- If so, whether is it more than six months ago. It can be indirectly converted into a **Divorce-Lawsuit** event (including *Sue Time*, *Initiator*, *Court*, *Sentence Time*, *Court Verdict* and *Result*). The argument *Time* can indirectly reflect whether it has been more than six months since the last time the divorce was filed.

Finally, we have got 13 types of events for divorce cases, as shown in Table 1. Here are the brief introductions of each event and its arguments (Each type of event has a general argument *Polarity* which is often an affirmative or negative word, and plays an important role in confliction detection):

● **Know**. The acquaintance event of the parties. The argument *Time* is usually a time expression. The argument *Participant* can only be parties.

● **Be-In-Love**. It means that the romantic relationship was established. The arguments are same as **Know** event.

● **Marry**. It means that the parties get married. The arguments are also the same as **Know** event.

● **Remarry**. This implies that the marriage between the parties is not the first time. The argument *Participant* can only be parties.

● **Be-Born**. A **Be-Born** event occurs whenever a child of parties is given birth to. If the child is still underage, raising questions should be taken into consideration, so we need to know the date of birth or age.

● **Family-Conflict**. Family conflicts reflect the status of mutual affection. Whether the relationship is broken or not is an important factor for the judge to decide whether to grant divorce. The argument *Participant* include not only the parties but also the entire family.

● **Domestic-Violence**. Divorce shall be granted if there is domestic violence circumstance. the argument *Participant* include not only the parties but also the entire family. The difference between **Family-Conflict** event and **Domestic-violence** event is that there are obvious abusers and victims in Domestic-violence event.

Table 1. Concerned Event Types of Divorce Dispute Cases

| Index | Event Type (*Abbreviation*) | Key Attribute | Non-Key Attribute |
|---|---|---|---|
| 1 | Know (*K*) | Time, Participant | Polarity |
| 2 | Be-In-Love (*BIL*) | Time, Participant | Polarity |
| 3 | Marry (*M*) | Time | Polarity |
| 4 | Remarry (*R*) | Participant | Polarity |
| 5 | Be-Born (*BB*) | Name | Time, Gender, Age, Polarity |
| 6 | Family-Conflict (*FC*) | Trigger Chunk | Polarity |
| 7 | Domestic-Violence (*DV*) | Time, Perpetrators, Victim | Polarity |
| 8 | Bad-Habit (*BH*) | Participant, Trigger Chunk | Polarity |
| 9 | Derailed (*DE*) | Time, Derailed-Person, Derailed-Target | Polarity |
| 10 | Separation (*S*) | Begin-Time, End-Time | Duration, Polarity |
| 11 | Divorce-Lawsuit (*DL*) | Sue-Time, Initiator | Court, Sentence-Time, Court-Verdict, Result, Polarity |
| 12 | Wealth (*W*) | Trigger Chunk | Value, Is-Common, Is-Personal, Whose, Polarity |
| 13 | Debt (*D*) | Debtor, Creditor | Value, Polarity |

● **Bad-Habit**. Divorce shall be granted if bad habits of gamble or drug addiction which remain incorrigible despite repeated admonition. we can't judge if remain incorrigible, so we just concern whether the parties have any bad habits, including alcohol, whoring, gambling, drug, pyramid selling, theft, fighting, net-addiction and fraud. The argument *Participant* should only be the parties.

● **Separation**. Divorce shall be granted if separation caused by incompatibility, which lasts two full years. We need to know the beginning time, the end time and the duration.

● **Divorce-Lawsuit**. Has either party filed for divorce before? And a lawsuit must be more than six months from the last lawsuit. So, the argument *Sue-Time* is the time when the **Divorce-Lawsuit** event is happened, the argument *Initiator* is the party who flies the lawsuit, the argument *Court* is the court where the case is heard, and the arguments *Result*, *Sentence-Time,* and *Court-Verdict* express the result of the case, the time of judgment and the referee documents.

● **Wealth**. At the time of divorce, the disposition of the property in the joint possession of husband and wife is subject to agreement between the two parties. In cases where an agreement cannot be reached, the people's court shall make a judgement. So, we concern at all property and their value, whether each property is community property or individual property.

● **Debt**. While divorcing, debts incurred by either husband or wife during their marriage shall be paid off with their jointly possessed property.

*3.3. Data Annotating*

For the pre-defined 13 types of divorce events, there is no structured database containing all the information items. So, we mark the litigation materials manually. We have marked 3100 case materials using the standard begin-inside-outside (BIO) scheme. The steps for labeling are as follows:

(1) **Pre-labeling**. 100 litigation materials are randomly selected for pre-labeling test, and then the judge judges whether the fragments we labeled belong to their concerns when handling divorce disputes cases. Adjust the labeling method based on the judge's feedback.

(2) **Annotation setting up**. We use the open source annotation tool BRAT[5], which is an online environment for text annotation. Examples of annotations are shown in Figure 4.

(3) **Annotator Training**. Training participants in the labeling. The training includes reviewing the facts in the trial process, the predefined 13 types of divorce disputes, and the use of marking tools.

(4) **Labeling**. Allocate enough time to the participants to ensure the accuracy of the labeling. There will be some cases where the choice of the appropriate trigger chunk is ambiguous. For example, Figure 4 E1. When labeling in Chinese, we should not only mark "sued ", but mark "sued for divorce from the defendant ". Because *sue* is not only limited to divorce dispute cases. The whole chunk "sued for divorce from the defendant" in Chinese is semantically complete.

In another example, shown as Figure 4 E2. The trigger chunk overlaps with the parameter gender. In this case, we just mark the trigger chunk, do not mark the overlapped argument.

Aiming at the phenomenon represented by the above examples, we propose two principles of annotation:

(1) The annotation content should be complete and contain all the information of each annotation item.

(2) In an event, when trigger words and arguments overlap, and trigger words have higher priority than arguments.

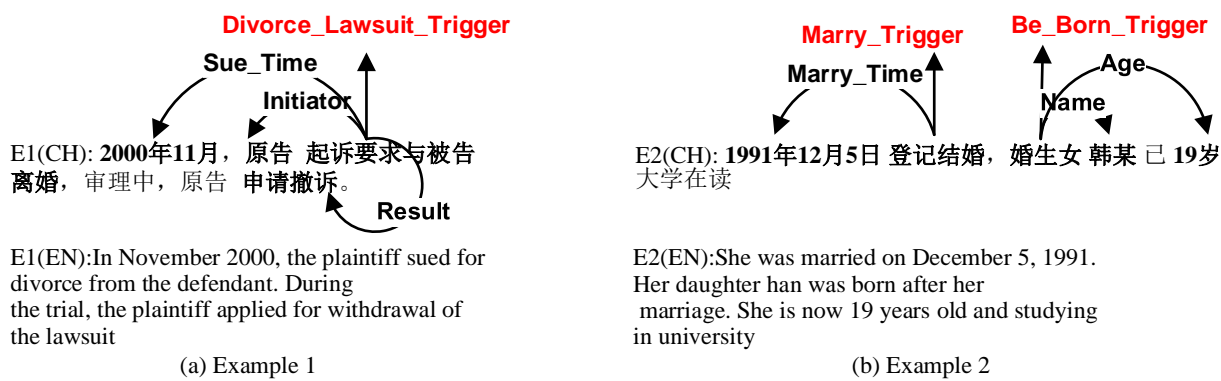

**Figure 4**. Two examples of annotation.

---
[5] http://brat.nlplab.org/about.html)

## 3.4. Event Extraction

Figure 5 shows the process of event extraction, which is described in detail in the rest of this subsection.

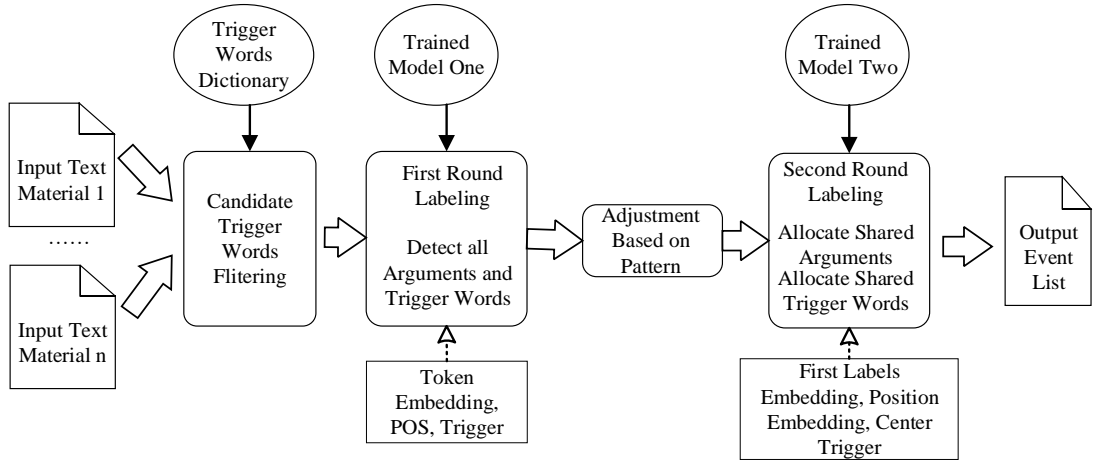

**Figure 5.** The Process of Event Extraction

**Trigger Words Dictionary.** For a specific cause of cases, we can collect as many trigger words as we can to form a dictionary, which can cover almost all trigger words of the predefined event types.

In the definition of ACE, an event's trigger is the chunk in its scope that most clearly expresses its occurrence. Events are usually triggered by verbs, nouns etc. But among our collection of trigger chunks, some are sentence fragments, which results in a very large number of trigger word chunks but they are similar to each other. Therefore, we compress and merge similar trigger words for each event type. We find out the similar trigger chunks for an event type, and extract the overlap between trigger chunks. And we take the shortest phrase which covers the overlap and has complete semantics as the final trigger chunk.

**Filtering Candidate Trigger Words.** For the input text, we first use LTP[6] to split into sentences. Then for each sentence, the trigger words dictionary is retrieved to determine whether it contains any trigger chunk. If so, the sentence may contain an event instance and a candidate trigger chunk have been determined. On the contrary, if it does not contain any trigger chunk, the sentence is almost impossible to contain the target events, so discard it.

**First Round Labeling.** The neural network has been confirmed to be able to automatically obtain the feature information of the text. And Bi-LSTM-CRF (i.e., Bidirectional Long Short-Term Memory followed by a Conditional Random Fields layer) has excellent performance in sequence labeling, such as named entity recognition (NER) and part-of-speech tagging (POS), in which Bi-LSTM are used to obtain the context feature, and CRF supplement the transfer information among the result labels to get the best labels sequence. Both Yang et al. (2018) and Zeng et al. (2018) use Bi-LSTM-CRF to extract events from a single sentence. In our approach, we also treat event extraction as sequence labeling.

As described in section 3.2.3, the target event types of divorce dispute contain a total of 13 types of trigger words, and 48 arguments. So, we can define 61 kinds of meaningful labels and a kind of meaningless label (O). To solve the problem of sharing arguments or trigger words, there are two ideas. The first idea is to set a new label type for each sharing situation, but the situations are complex and diverse. And the number of labels will increase dramatically, which will lead to the problem of sparse data. The second idea is to make multiple predictions for each sentence, each of which only takes a trigger chunk and the corresponding event into consideration. But this will lead to the number of meaningless label (O) far more than the number of meaningful labels. Finally, we adopt the two-round-labeling method, which combines the two ideas.

**Table 2.** Transition Labels and the Meaning

| Transition Label | Meaning | Transition Label | Meaning |
|---|---|---|---|
| Time | Time | Is-Common | Whether it belongs to common property |
| Person | Participant | Money | Property value |
| Name | Name of the child | Court | Court that accepts the divorce case |
| Gender | Gender of the child | Document | Court verdict |
| Age | Age of the child | Result | Judgment result of divorce case |
| Duration | Separation duration | Polarity | Affirmative or negative word |
| Is-Personal | Whether it is personal property | | |

---

[6] NLP tool belongs to Harbin Institute of Technology, http://www.ltp-cloud.com/

Statistics show that the arguments shared by multiple events are usually *Participant*, *Time* and *Polarity*. The rest arguments are unique to an event (such as arguments *Name*, *Gender* and *Age* for the **Be-Born** event, the *Court-Verdict* for the **Divorce-Lawsuit** event, etc.). So instead of directly predicting the final label for each token, we define 13 transition labels (Table 2) in conjunction with shareable arguments and event-specific arguments. The original 48 kinds of arguments are mapped to 13 transition labels, reducing the label categories and increasing the frequency of individual labels. In the first round labeling, we just need to identify transition labels.

The structure of model for predicted transition labels is shown in Figure 6. The annotated data would be converted into begin-inside-outside (BIO) scheme to train the model. When a word is used as the beginning of an argument or a trigger, it is marked as B_Label; if a word is not the beginning of the argument, it is marked as I_Label; the other is labeled O_Label.

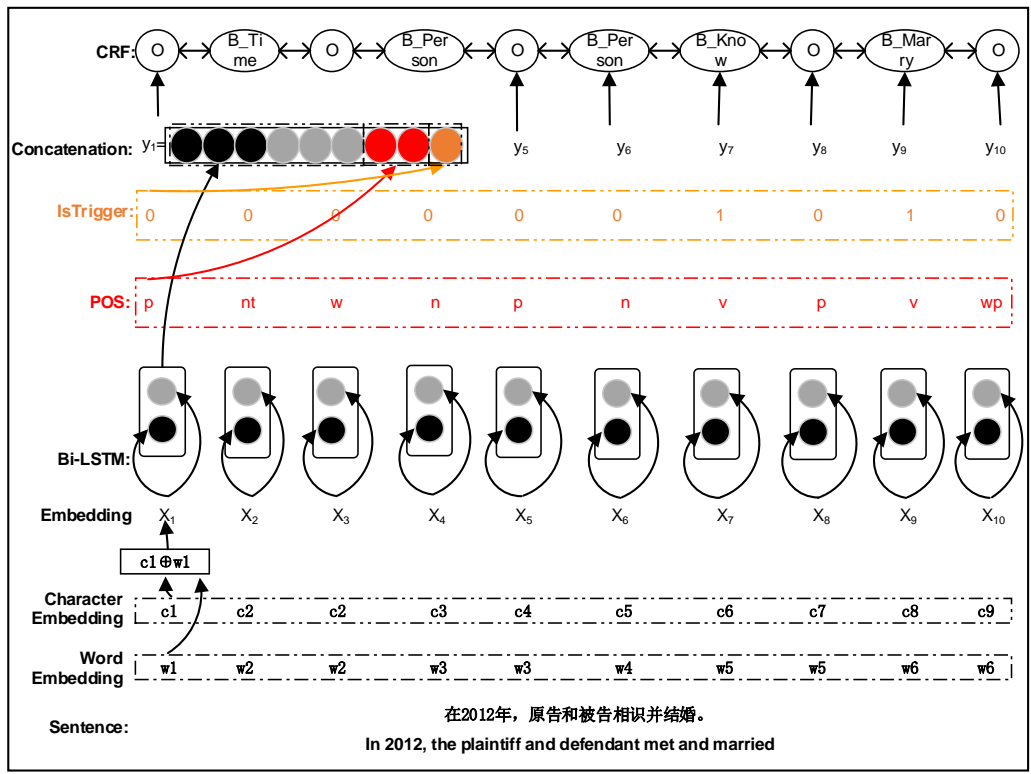

**Figure 6**. The Architecture of First Round Labeling

Introductions of the process for predicting transition labels and the core layers shown in Figure 6 are as following:

● **Pretreatment**. There is no interval between Chinese, so we first use the tool PYLTP of Harbin Institute of Technology to segment words. Set the maximum length of a sentence as MAX_SEQUENCE_LENGTH, if the number of words in a single sentence is less than the maximum, then pad the tag <PAD> in the right up to MAX_SEQUENCE_LANGTH; and cut if more than the length. We set MAX_SEQUENCE_LENGTH to 55, which can exceed 97% of the actual sentence length.

● **Text Representation**. Text representation is the most vital work in NLP tasks. Its goal is to transform a varied-length text into a low-dimension vector and the distance of vectors indicates the similarity of components, i.e., embedding (Wu, Zhao, & Li, 2020). The most classical vectorization technology is Word Embedding which finds a mapping relation between word space and continuous vectors space. It expresses the relevance of words by the distance between vectors. Unlike English, Chinese characters also have some semantic information in addition to words. The meaning of the character itself and the meaning of the word often complement each other, so we combine the character vector with the word vector to obtain more complete semantic information. In this paper, two models for generating a 300-dimensional vector for Chinese character and a 300-dimensional vector for Chinese word are trained according to Word2vec[7] based on 40,000 civil judicial documents downloaded from Chinese Judgment Online[8] using Skip-Gram-Model. We concatenate the vector representation of each character and the vector representation of the word that the character is in.

● **Bi-LSTM (**Bidirectional Long Short-Term Memory**)**. The Recurrent Neural Network (RNN) can remember the historical semantic environment by hiding the state, but there is the problem of gradient disappearance or explosion. Long Short-Term Memory (LSTM) neural networks can solve the problem through the input and output of gated information. We use Bidirectional-Long-Short-Term-Memory neural networks in this paper, combining the semantic information of the words before and after.

---

[7] Gensim Word2vec, https://radimrehurek.com/gensim/models/word2vec.html
[8] https://wenshu.court.gov.cn/

● **Concatenation**. Word Encoding (Semantic features of word) obtained by Bi-LSTM, POS Embedding (POS tags can well represent the syntactic characteristics in natural language statements), and Whether each word belongs to a candidate trigger chunk (determined in section 3.4.2 and we use 0 and 1 to indicate). The dimension of the result vector after connection shown in Figure 7 is not the real dimension set in the experiment.

● **CRF**. LSTM can find a tagging result with the maximum probability more from the perspective of the semantics of a single word, but it ignores some restrictions about tags, such as tags must start with B_Label, and following tags of the same class must be I_Label. Conditional Random Fields (CRF) is a popular probabilistic method for structured prediction (Sutton and McCallum, 2012). So, we add a layer of CRF to get transfer information between tags.

● **Output**. The output is the ID sequence of predefined transition labels, corresponding to 27 kinds of meaningful types labels (including 13 kinds of trigger chunk types, 13 kinds of transition labels of parameters and others label O). Outside of the others label, there are all are 52 kinds in total considering B_Lable and I_Label.

**Adjustment Based on Pattern.** In the experiment we found that although we use transition labels to reduce the problem of data sparsity, the two labels, *Polarity* and *Money,* are still relatively few, which lead to low accuracy. But the two labels have strong regularity. So, we optimize the prediction results of the first round based on pattern:

> Label *Polarity*. We collect positive words and negative words as much as possible, constructed a polarity dictionary. Then we retrieve and label the polarity words in a sentence other than the result of the first round labeling.
>
> Label *Money*. We use a regular match to get the word chunks that involve numbers in a sentence, and label them.

**Second Round Labeling.** So far, we obtain all trigger chunks and transition labels, which represent all meaningful labels. We then assign the shared trigger chunks and arguments to different events mentions. We use different allocation strategies for shared arguments and shared trigger chunks:

**(1) Multi-events share parameters in single sentence**. The arguments are allocated multiple times to different events mentions. But the unique arguments are assigned once. For each sentence, we use a CRF model to predict the final labels of the transition labels, in which we only consider one trigger chunk once time. If there are several trigger chunks in a single sentence, it is necessary to assign the result of the first round several times and obtain a complete event instance each time. The structure of model for assigning shared parameters is shown in Figure 7.

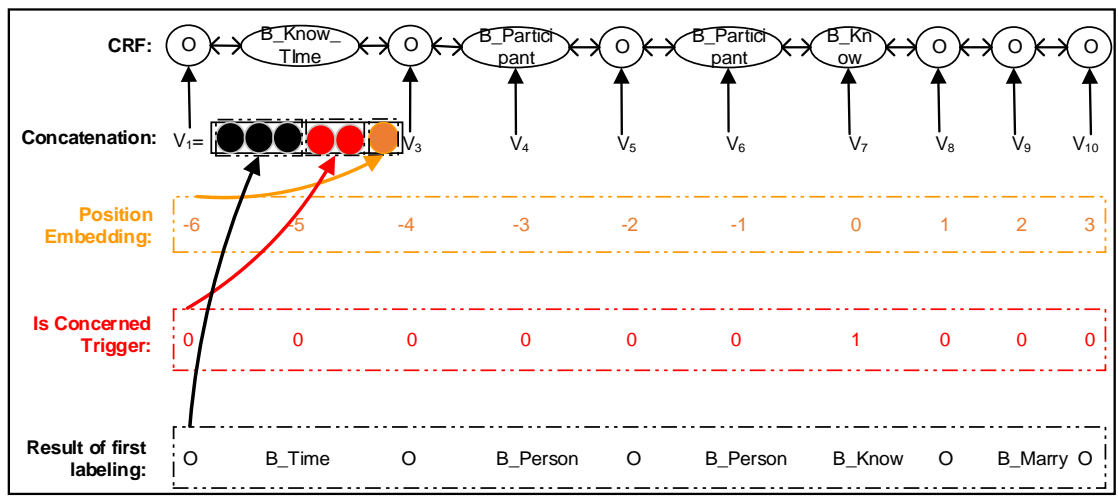

Figure 7. The Structure for Assigning Shared Parameters in Second Stage

The core differences between the first round and the second labeling are Concatenation and Output layers:

● **Concatenation**. (1) Result of first labeling: the embedding of transition labels and trigger labels. (2) One hot expressions of labels are used to represent the 53 types of labels generated in the first stage. (3) Whether the word belongs to the trigger chunk of currently concerned. If it is, we mark it with 1, otherwise 0. For example, two samples would be generated for the instance in Figure 6 and they are 0 0 0 0 0 1 0 0 0 (taking B_Know as the concerned trigger chunk) and 0 0 0 0 0 0 0 1 0 (taking B_Marry as the concerned trigger chunk). (4) Position Embedding: the embedding of the position relative to the center trigger chunk. The relative position of the word before the trigger block is negative, the word belonging to the trigger block is 0, and the word after the trigger block is positive.

● **Output**. The output is the label types of the final labels for 13 kinds of trigger types, 36 kinds of arguments types and a kind of meaningless label (O). One sample of output represents an event.

**(2) Multi-events share trigger chunk in single sentence**. Statistics shows that the phenomenon of sharing trigger chunks mainly occurs in the **Be-Born** event and **Divorce-Lawsuit** event. For **Be-Born** event, only argument *Time* can be shared, and the rest arguments are unique, such as arguments *Name*, *Gender* and *Age*. It can be defined as the following rule:

*If in a result sequence of first round, the number of a kind of unique parameters of **Be-Born** event including Name, Gender and Age are more than the number of the trigger chunk of **Be-Born** event, the number of **Be-Born** events instances is equal to the maximum number of unique arguments. Trigger chunk and other arguments are shared by multiple **Be-Born** events.*

For the example shown in Figure 3, the sequence of first round should be:

O B_Be_Born B_Gender B_Name O B_Gender B_Name

The number of *Gender* and *Name* both are 2, and the number of trigger chunk is 1. We can infer that there two **Be-Born** events instances and the trigger chunk is shared. Similarly, for **Divorce-Lawsuit** event, following rules can be defined:

(1) *For unique arguments of **Divorce-Lawsuit** event (including Court, Document, Result), if a type of unique arguments has values more than the trigger chunk, the number of **Divorce-Lawsuit** events instances is equal to the maximum number of unique arguments, and trigger chunk and other arguments are shared by multiple "Divorce-Lawsuit" events instances.*

(2) *For the argument Time in **Divorce-Lawsuit** event, if there are arguments Time that are not parameter value of other events. The number of **Divorce-Lawsuit** events instances is equal to the number of arguments Time on the same side of the trigger chunk. The trigger chunk and other arguments are shared.*

The problem of sharing trigger chunk can be solved by the above rules.

*3.5. Event Alignment*

Currently, events are extracted from the statement of the plaintiff and the defendant, which is not enough, because conflicts are inevitable among these extracted events and these conflicts are helpful for judges to classify the disputed issues between the parties. So, we try to find the views of plaintiff and defendant on the same event instance, and determine whether the two views conflict. We first align the events that are co-reference. And then we check whether the events are conflict. In this section we describe how to align events.

Among the predefined 13 types of events, some types are unique, which can happen only once in the marriage life of the parties. And we call them **Unique-Event**. **Unique-Event**s are automatically aligned, including **Know** and **Be-In-Love** events. Each of the other 11 types of events, named **Non-Unique-Event,** may have multiple instances. We define the attributes of **Non-Unique-Events** that can identify the event instance uniquely as key attributes (possibly including trigger words). And other attributes as non-key attributes. (Shown in Table 1) Two events which belong to the same type of **Uon-Unique-Event** are co-reference if all key attributes are equaled.

When determining whether the values of the same attribute of two events are equal, we follow the rules:
(1) *If either of them does not exist, they are equal.*
(2) *For argument Time, we get the exact date from the time expression and compare the Year, Month and Day respectively.*
(3) *For the trigger chunks of **Family-Conflict** and **Wealth**, we calculate similarity between text chunks (text editing distance divided by the maximum text length of two text chunks), and judge whether they are equal or not by comparing the similarity with a predefined threshold. Through experimental statistics, we finally set the threshold value for **Family-Conflict** trigger chunk as 0.5, and the threshold value for **Wealth** trigger chunk as 0.75.*
(4) *For argument Participant, we treat them differently according to the scope of the participant:*
　(4.1) *Participant can only belong to both parties, such as **Marry**-Participant, **Bad-Habit**-Participant, **Derailed**-Person, and Initiator. We normalize the values to "plaintiff" or "defendant". If the extracted text chunk contains "plaintiff" or "defendant", we set the value as the same If no, combine the pronouns "I", "she/he" and the person who speak the statement to normalize.*
　(4.2) *Participant not only belongs to parties, such as Perpetrators, Victim, Debtor and Creditor. If pronouns such as "I" or "She/He" are included, converted to plaintiff or defendant according to the speaker. Otherwise, the extracted text chunks are directly compared whether are equal.*
(5) *For the trigger chunk of **Bad-Habit**, we know the bad habit categories (see section 2.1.3), so if the bad habit categories are the same, they are equal.*
(6) *For **Separation** events, they are aligned if there is overlap between the duration of two events from the*

*beginning to the end.*

(7) *The rest directly compares two text chunks to see if they are same.*

## 3.6. Event Confliction Detection

In the previous section, we defined **Unique-Event** and **Non-Unique-Event**. The Unique-Events are automatically aligned and we align the **Non-Unique-Events** according to their key attributes. In this section, we will perform confliction detection for events. We stipulate that two events must meet the following preconditions for further confliction detection:

(1) *Belonging to the same type of events.*

(2) *Belonging to **Unique-Event**, or belonging to **Non-Unique-Event** and aligned.*

For two events that meet the two conditions, if all attributes of the two events are equal, they do not conflict. Otherwise, they conflict. So, for the 13 types of events, we can determine if there is a confliction by following rules:

(1) **Know** *event belongs to **Unique-Event**. First, judge whether the values of argument Time of two events are equal, and then judge whether the values of argument Polarity are the same semantics (positive/negative). We construct a polarity dictionary that contains positive and negative words and judge polarity by retrieving the dictionary.*

(2) **Be-In-Love** *event belongs to **Unique-Event**. First, judge whether the values of argument Time of two events are equal, and then judge whether the values of argument Polarity are the same semantics (positive/negative semantics).*

(3) **Marry** *event belongs to **Non-Unique-Event**. After alignment, judge whether the values of argument Polarity have the same semantics.*

(4) **Remarry** *event belongs to **Non-Unique-Event**. After alignment, judge whether the values of argument Polarity have the same semantics and whether trigger chunks have the same meaning (first marriage or remarriage). We construct a dictionary that contains first marriage and remarriage words and judge the meaning of trigger chunks by searching the dictionary.*

(5) **Be-Born** *event belongs to **Non-Unique-Event**. After alignment, judge whether the values of arguments Time, Gender, and Age are equal. If equal, judge whether the values of argument Polarity have the same semantics.*

(6) **Family-Conflict** *event belongs to **Non-Unique-Event**. Firstly, judge whether the trigger chunk have the same meaning. Most of the time, **Family-Conflict** events are described negative state of the relationship between two parties, and a few are positive. So, we collect the trigger word set for positive emotional states. After alignment, the positive emotional trigger words set is retrieved to determine whether the values of trigger chunks are positive or negative. If they are the same emotional tendency, they are considered equal, and then judge whether the values of argument Polarity have the same semantics.*

(7) **Domestic-Violence** *event belongs to **Non-Unique-Event**. After alignment, judge whether the values of argument Polarity have the same semantics.*

(8) **Bad-Habit** *event belongs to **Non-Unique-Event**. After alignment, judge whether the values of argument Polarity have the same semantics.*

(9) **Derailed** *event belongs to **Non-Unique-Event**. After alignment, judge whether the values of argument Polarity have the same semantics.*

(10) **Separation** *event belongs to **Non-Unique-Event**. After alignment, if the values of arguments Begin-Time and End-Time are not equal, it is considered as conflict. If they are equal, judge whether the values of argument Polarity have the same semantics.*

(11) **Divorce-Lawsuit** *event belongs to **Non-Unique-Event**. After alignment, it shall judge whether the values of argument Court, Sentence-Time, Court-Verdict and Result are all equal. If equal, check if the values of argument Polarity have the same semantics.*

(12) **Wealth** *event belongs to **Non-Unique-Event**. After alignment, two wealth events conflict if one party considers the wealth as common property while the other considers it as personal, or both parties consider it as personal but belongs to different parties, or the values of arguments Values or Polarity are different.*

(13) **Debt** *event belongs to **Non-Unique-Event**. After alignment, judge whether the values of argument Value are equal. If equal, then check if the values of argument Polarity have the same semantics.*

Through the above principles, we check the extracted events and show them to help the judges quickly understand the case. And detect whether there is a conflict in these events, and highlight if there is a conflict, as a reference for the judge to find the disputed issues. Besides, we define pairs of events which are aligned but not conflicting events to be ***entailment***.

## 4. Evaluation

*4.1. Data Preparation*

We have annotated litigation textual materials of 3100 divorce cases. Each case has two types of documents, i.e., statements of the plaintiff and the defendant respectively. The numbers of each type of event are shown in Figure 8(a). While building the event extractor, each document is viewed as a single item of the data set. We divided documents in the data set into three intervals according to document size, namely [0,3) KB, [3,6) KB and [6,+∞) KB. As mentioned in section 2.3, our approach is designed to handle the challenge of multi events sharing the same trigger words or attributes. The statistics of multi events that sharing trigger words or attributes of dataset we prepared is as shown in Figure 8(b). Only those with a number greater than 10 are presented in the figure. We can see that **_Know_** event and **_Be-In-Love_** event share *Time* argument for 481 times, and two **_Be-Born_** events share the trigger chunk for 124 times.

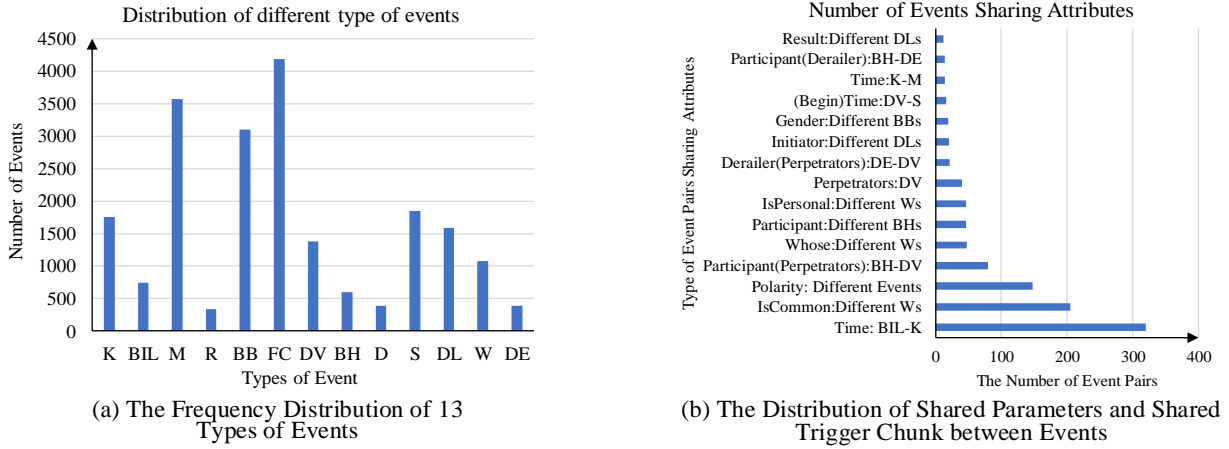

(a) The Frequency Distribution of 13 Types of Events

(b) The Distribution of Shared Parameters and Shared Trigger Chunk between Events

**Figure 8**. Statistics on the experimental data sets: distributions of event types and event pairs that sharing attributes.

*4.2. Effectiveness of the Proposed Approach*

**4.2.1 Evaluation Metrics.**

We defined the precision $P_i$, recall rate $R_i$ and $F1_i$ to evaluate the effectiveness of predicting a single label, as follows:

$$P_i = \frac{S_i}{S_{ti}}, R_i = \frac{S_i}{S_{pi}}, F1_i = \frac{2P_iR_i}{P_i + R_i}$$

$S_i$ is the number of *i*-th label which are predicted correctly. $S_{ti}$ is the actual number of *i*-th label. $S_{pi}$ is the number of predicted *i*-th label. We define the P, R and F1 for all label showing in Table 2. The O label is not included in the evaluating objection. Since the distribution of different labels is not balanced over the data set, we use both Micro and Macro average of P, R, and F1 values of all labels for comparing feature functionality in feature ablation. To evaluate the final effect of event extraction, we define precision $P_{ee}$, recall $R_{ee}$ and $F1_{ee}$ as follows:

$$P_{ee} = \frac{S}{S_t}, R_{ee} = \frac{S}{S_p}, F1_{ee} = \frac{2P_{ee}R_{ee}}{P_{ee} + R_{ee}}$$

$S_t$ is the total number of all trigger chunks and arguments chunks, S is the total number of the correct trigger chunks and arguments chunks in the result of event extraction. $S_p$ is the total number of trigger chunks and arguments chunks in the result of event extraction.

The foundation of detecting conflicts among events accurately is to align event pairs correctly. For evaluating performance of event alignment, we define aligning precision and recall values ($E_{aligned}$ is the set of aligned event pairs and $E_{should\ aligned}$ is the set of event pairs that should be aligned.):

$$P_{align} = \frac{|E_{aligned} \cap E_{should\ aligned}|}{|E_{aligned}|}, R_{align} = \frac{|E_{aligned} \cap E_{should\ aligned}|}{|E_{should\ aligned}|}$$

However, the ultimate goal of the proposed approach is to detect conflicts among events, which means the performance of detecting results is more explicit in telling the effectiveness. The precision and recall of detecting conflicts are defined as following ($E_{contradictory}$ is the set of predicted conflicting event pairs and $E_{should\ contradictory}$ is the set of event pairs that are truly conflicts.):

$$P_{contradictory} = \frac{|E_{condradictory} \cap E_{should\ contradictory}|}{|E_{contradictory}|}, R_{contradictory}$$

$$= \frac{|E_{condradictory} \cap E_{should\ contradictory}|}{|E_{should\ contradictory}|}$$

### 4.2.2 Experiment Settings

***Data splitting***: Each interval of the data set is randomly divided into ten pieces. Then, to make a 10-fold data set for cross validation, we randomly select and remove one piece from each of the three intervals to combine one-fold. This was repeated for ten times. Eventually, each fold consists of the same size distribution of texts. However, we do not restrict the same distribution of event types over all folds. In cross validation, each fold was used as test set and validation set once. The average result of 10-folds is viewed as a single result of the proposed approach.

***Environment and parameter selecting:*** The adopted deep learning framework is Tensorflow[9]. An NVIDIA V100 TENSOR CORE GPU (32GB) is used for training the deep learning model. The max-length of the input is set to 55 Chinese words. The batch size is set to 64. The number of epochs is tuned to maximize accuracy on the development data using grid search. We tried 10, 15 and 20.

### 4.2.3 Event Extraction: Compare with Baselines.

There are three baseline methods[10].

***Directly Predict***: Different from the proposed two-round-labeling strategy to obtain the final label sequence, this method directly takes trigger chunk as the center and directly predicts the final labels of a single sentence, which will be conducted as many times as the number of trigger chunks in the sentence.

***Zeng et al. (2018)***: There are many methods for event extraction, we will introduce them in the next section. But none other than Zeng et al. solves the sharing problem. So, we compared our approach with this which uses Bi-LSTM-CRF to obtain the probability distribution of possible labels of each word, and then an ILP-solver is used to output multiple optimal sequences, corresponding to multiple events.

***Yang et al. (2018)***: For better comparison, we add another method, proposed by Yang et al., as the other baseline, which does not consider single sentence containing multiple events. They use character embedding as input and the uses Bi-LSTM-CRF to extract events from a single sentence without considering the case of sharing. So, we set the sharing situation to a new label in the contrast experiment. We pick the top 8 shared situation and add 8 kinds of new label to replay experiment of Yang et al. (2018).

Table 3: Compare our Two-round method with baselines.

| Approach | $P_{ee}$ | $R_{ee}$ | $F1_{ee}$ |
|---|---|---|---|
| Directly Predict | 0.66 | 0.74 | 0.70 |
| Zang et al. (2018) | 0.64 | 0.71 | 0.67 |
| Yang et al. (2018) | 0.68 | 0.68 | 0.68 |
| Ours | **0.72** | **0.78** | **0.75** |

**Results and Discussion**. We conducted the 10-fold cross validation of our method for 12 times, as well as the two baselines. Paired T-test results show that our approach is significantly outperforming the baselines. Table 3 shows the experimental results. As we can see, the F1 of two-level labeling method is 0.05 higher than that of direct prediction. We speculate that the reason is the direct prediction of the final label lead to there are too many kinds of labels. And compared with meaningless label O, the number of meaningful arguments labels are smaller, resulting in the problem of sparse data. And the method of Zeng et al. (2018) use an ILP-solver to output multiple events, so its recall rate is higher than Yang et al. (2018). But it also leads to its low accuracy. Both the precision and recall rate of our method are better than those two methods.

### 4.2.4 Ablation on different stages of Event Extraction.

In order to compare the influence of each date processing step on the result, the feature ablation experiments of each round are conducted. Table 4 shows the results.

***First Round Labeling***: POS, Trigger.

In the first stage, in addition to semantic features, we also introduce POS features and trigger-chunk features. Therefore, we conducted experiments where POS and trigger-chunk features are not adopted. Table 4 shows the results. Comparing both Micro and Macro average Precision, Recall and F1 values of all experiments, we find that they almost have the same increasing or decreasing performances. This means that, from the overall point of view, the functionality of features is stable and would not influenced by the unbalanced distribution of different labels.

---

[9] https://www.tensorflow.org/
[10] Source code of baselines and the proposed approach can be found here: https://github.com/shengxiaoyu/baseline.git and https://github.com/shengxiaoyu/BiLSTM_CRF_EVENT_DETECT.git

Table 4: Experimental results of feature ablation in different labeling rounds

| Approach Stage | Identifier | Feature Type | Average Precision | | Average Recall | | Average F1 | |
|---|---|---|---|---|---|---|---|---|
| | | | *Micro* | *Macro* | *Micro* | *Macro* | *Micro* | *Macro* |
| First Round Labeling | FR1 | -POS, -Trigger | 0.75 | 0.64 | 0.60 | 0.45 | 0.67 | 0.50 |
| | FR2 | +POS, -Trigger | 0.79 | 0.66 | 0.65 | 0.46 | 0.71 | 0.51 |
| | FR3 | -POS, +Trigger | 0.84 | **0.78** | 0.84 | 0.64 | 0.84 | 0.68 |
| | FR4 | +POS, +Trigger | **0.86** | 0.77 | **0.85** | **0.66** | **0.85** | **0.69** |
| Second Round Labeling | SR1 | -Position | 0.92 | 0.76 | 0.70 | 0.51 | 0.80 | 0.58 |
| | SR2 | +Position | 0.96 | 0.81 | **0.80** | **0.60** | **0.87** | **0.65** |
| | SR3 | Rule-based Matching | **0.98** | **0.86** | 0.60 | 0.48 | 0.74 | 0.56 |

Comparing the FR2 experiment and FR1, it can be found that the POS tag of words has a positive influence in predicting their labels, since it helps increase the micro and macro average precision/recall by 0.04/0.05 and 0.02/0.01 respectively. As we surmise that POS tags represent the syntactic characteristics in natural language statements well. Besides, by reading raw data, we found that the parameters and trigger of events often have relatively fixed grammatical roles in a sentence. Comparing the FR3 line with FR1, it can be found that if adopt the trigger word feature only, the micro and macro average F1 values can be improved by 0.17 and 0.18 respectively. This is mainly because that the candidate trigger word derived from the trigger word dictionary is likely to be the trigger word of an event, which greatly improves the prediction accuracy of the trigger word and related parameters. Eventually, we apply the two features together in experiment FR4, the performances are improved slightly again.

Besides, we found that the improvement made by the Trigger words is much higher than that made by the POS tag. This is largely because that the trigger words represents the explicit understanding of human beings of the natural language, which are difficult for the model to learn, and they have much closer connection with the event definition, i.e., extracting goals, than the POS tag. This also tells that while choosing features for machine learning algorithms, the features of strong connection with learning goals from the human being's understanding perspective are better, i.e., telling the model extra related human knowledge is a good choice to improve the performance.

***Second Round Labeling***: Position Embedding, Rule-based Matching.

In the second stage, to solve the problem of multi-events sharing parameters in single sentence, we take a trigger chunk as center and assign the transition labels multiple times. But in addition to the center trigger chunk and the transition labels, we also introduced position features, which represents the relative position of each word and center trigger chunk. We think this information is useful for redistribution of transition labels, especially when there are multiple transition labels of the same type. We compared the effects before and after the introduction of the position features. Experimental results shown in Table 4 indicates that adding the feature vector of position features can significantly improve the prediction result of the second stage.

After getting the results of transition labels in the first stage, we can also just allocate the transition labels based on the rules of event arguments distribution instead of using the sequence labeling method CRF. We can define some rules based on statistics. For example:

● *The shared transition labels of Know, Be-In-Love and Marry events include Person and Time, and we can find the nearest Person and Time transition arguments before the event trigger chunk which are taken as the arguments of the events.*

● *Remarry event sharing transition arguments include Person, and the nearest Person type argument before the event trigger chunk is taken as the event argument.*

In combination with these rules and the rules for solving multi-event which share trigger chunk in a single sentence, the second stage can use a completely Rule-based Matching approach. The results of rule-based matching approach shown in Table 4 prove that model predict is 0.13 higher than rule-based allocation.

### 4.2.5 Effectiveness of Event Alignment and Conflict Detection

As for event conflict detection, we detect whether any two events in the test set can be aligned, and detect whether the aligned events are conflicting to each other. The confusion matrix of aligning and conflict detecting results is shown in Table 5. Events that should not be aligned are marked as **Non**. Predicting results of each type of events are marked with a prefix **Predicted**. According to the results, the precision of aligning approach, i.e., $P_{align}$, is 0.296, and the recall, i.e., $R_{align}$, is 0.891. The precision and recall of conflict detecting, i.e., $P_{contradictory}$ and $R_{contradictory}$, are 0.811 and 0.803 respectively. Since there are many events that should not be aligned combining with another event to make alignment pairs, the precision of aligning is very low. However, most of these wrongly aligned event of type **Non** are predicted as **Entailment**, which means the precision of predicting conflicts would not be influenced by the low precision of aligning. The high recall of aligning, which is 0.891, is the foundation of high recall of predicting conflicts, because conflicts detecting would only be performed in aligned event pairs. However, in all aligned event pairs, if there are many contradictory event pairs predicted as **Entailment**, the performance of conflict detecting would also very low. So, it is necessary to calculate the precision and recall of conflict detecting. Table 5 shows that both the precision and recall of conflict detecting are higher than 0.8, which means most of the conflicts in cases are detected.

There are two aspects to improve the performance of conflict detecting. First, improving the performance of aligning, i.e., distinguish event pairs of **Non** but predicted as **Contradictory** from true contradictory pairs, as well as distinguish event pairs of **Contradictory** but predicted as **Non** from truly should not be aligned pairs. Second, improving the accuracy of classifying **Contradictory** and **Entailment** pairs. According to Table 5, the second aspect for improving plays a more important role in enhancing conflict detecting. This would be a future work of our research.

Table 5: The Results of Event Conflict Detection

|  | **Predicted-Contradictory** | **Predicted-Entailment** | **Predicted-Non** |  |
|---|---|---|---|---|
| **Contradictory** | 159 | 26 | 13 | $R_{contradictory}=0.803$ |
| **Entailment** | 22 | 216 | 39 | $R_{align}=0.891$ |
| **Non** | 15 | 992 | 71932 |  |
|  | $P_{contradictory}=0.811$ | $P_{align}=0.296$ |  |  |

*4.3 Effectiveness of JIA System (i.e., Judicial Intelligent Assistant)*

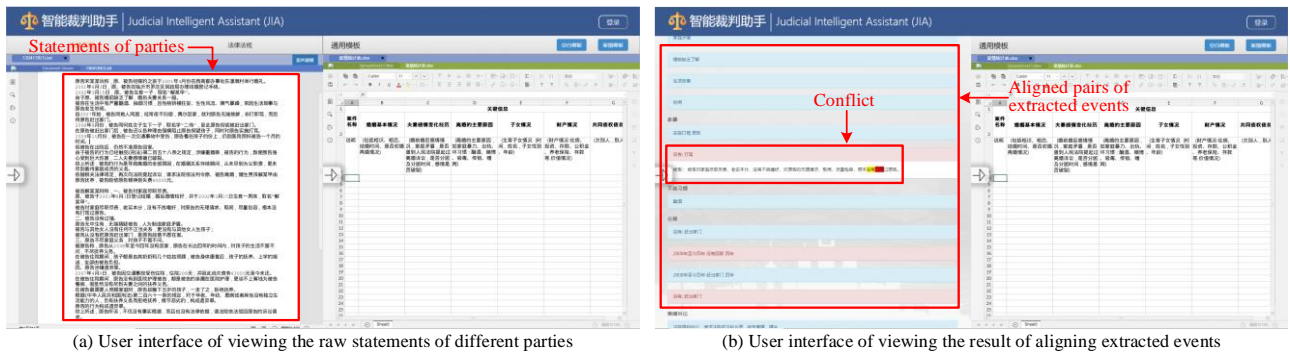

(a) User interface of viewing the raw statements of different parties    (b) User interface of viewing the result of aligning extracted events

**Figure 9**. Screen shots of the Judicial Intelligent Assistant system.

Based on the above steps, we implement the system that extract the target events from the textual litigation materials, then detect and highlight conflicting events. The target events can be extracted by uploading the litigation material containing the statements of case. Figure 9(a) and (b) show screenshots of the system that viewing the litigation documents and the event extraction results.

**4.3.1 Experiments Design**

Referring to the "*The Guide of Divorce Dispute Trial*" which are compiled by the judges in actual work, we determined 6 types of key information which consists of 13 predefined events, 4 types of possible dispute issues which could be got from the extracted events. For example, one type of key information is the basic situation of the marriage, such as the time of marriage, children, etc. Another is the main reason of forming the divorce dispute, such as domestic violence, derailment, etc. Possible issues of dispute include whether the relationship is broken down, whether a deal can be reached on property determinations, whether the creditor's rights and debts of the two parties are consistent, and so on. We choose 40 cases from the dataset for evaluating the JIA system. The selecting criterial is the number of words in the document since it costs different time for reading different lengths of documents. There are 10 cases for four different scales of words respectively, i.e., 300±30, 600±30, 900±30, and 1200±30.

For comparing the effectiveness of JIA for different people, we invite two groups of testers, i.e., 8 judge (Experts) and 8 normal persons (Novice, without specific judicial knowledge), to conduct the following procedure at the same time:

(1) Learn the "*The Guide of Divorce Dispute Trial*" to understand the 4 types of dispute issues and 13 types of events.

(2) Choose another 10 cases randomly from initial dataset for training testers. The training included extracting key information based on raw materials and clarifying the dispute focus, and determining the key information and dispute focus based on the system extraction result.

(3) The 8 testers were divided into two sub-groups A and B. Group A extract result based on raw materials, and Group B extract based on JIA system. All 8 testers begin to process the material at the same time, and record the time spent by each tester until complete the task.

**4.3.2 Efficiency Evaluation**

The statistics of sub-groups of both Expert-group and Novice-group are shown in Table 6. As we can see, for the same cases, whether the Expert-group or the Novice-group, if the target information is extracted based only on the raw material, the time increases as the text length of the raw material. By using the JIA system, both the Expert-group and the Novice-group do not have a positive correlation with text length in general, which are basically stable. We can draw the preliminary conclusion that the system can help the tester to reduce the time to understand the cases and obtain the dispute issues.

Table 6: The Time Spent by Groups A and B in the Expert-group

| Tester Type | Material Type | 300 Words | 600 Words | 900 Words | 1200 Words |
|---|---|---|---|---|---|
| Experts | Raw | 12m42s | 14m24s | 22m48s | 29m14s |
|  | JIA System | 8m34s | 7m44s | 10m34s | 8m12s |
| Novice | Raw | 44m10s | 64m10s | 96m40s | 113m50s |
|  | JIA System | 20m23s | 15m15s | 28m15s | 29m55s |

**4.3.3 Effectiveness Evaluation**

In addition, we make a deeper analysis of the events extracted by the Expert-group, as is shown in Figure 10 (a) and (b). For events belong to 6 types of key information, most of them (76.49%) can both acquired from system and raw materials. And 16.33% usually can only get from raw materials. 7.17% of events, which are lose when get directly from the raw materials, are got exactly with the system, i.e., Figure 10(a). For events (76.49%) that can both be obtained using either the system or raw materials, we calculated the actual number of event parameters and event trigger words is 872. And the coverage of system extraction results is shown in Figure 10(b). For the extracted event, we can get the basic complete event details.

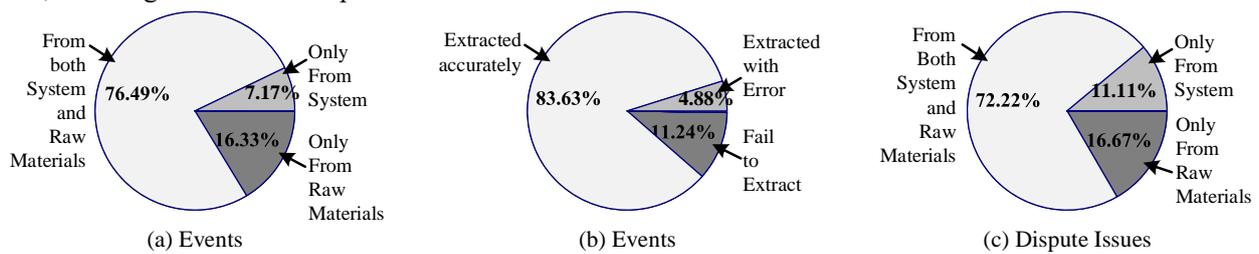

(a) Events     (b) Events     (c) Dispute Issues

**Figure 10**. The Statistics of events extracted From System or raw Materials by the Expert-group

As is shown in Figure 10(c), for the 4 types of possible dispute issues, our system can get 83.33% (72.22%+11.11%), and 16.67% cannot acquire only by the extracted result of system. There are two reasons. The first and main reason is error propagation. Some events (16.33%) cannot be obtained accurately in the event extraction phase on which the confliction detection based is affected. The second reason is that a few dispute issues required some semantic inference other than the extracted events. In one of the experimental cases, the plaintiff file a claim that the two parties split up one of self-built houses where they now live. The defendant confirms the existence of such a house, but proposes that the money for the construction and decoration of the house comes from the defendant's parents, so he did not agree with the partition. Here we can extract such a property of the house, but we can't tell from the explicit words that the defendant thinks it is not a joint property, because of which there is a disputed issue on the division of the house. Nevertheless, we can still believe that our method is helpful in identifying the focus of the dispute issues.

## 5. Related Work

**Artificial Intelligence and Law**: The history of Artificial Intelligence (AI) and Law started about four decades ago. Law is a rich test bed and important application field for logic-based AI research (Prakken & Sartor, 2015). For most of its existence, this discipline has been primarily concerned with attempts to formalize legal reasoning and create expert systems (Contissa et al., 2018). Recently, research on AI applications in legal domain has slightly shifted from developing rule-based, argument-based, and case-based models for representing legal knowledge and reasoning to adopting machine learning and text analytics to accomplish tasks in legal domain effectiveness and efficiently. There are a lot of NLP-related (Natural Language Processing) applications listed in (Contissa et al., 2018). But there are also applications applying computer vision (Sajjad et al., 2019), pattern detection (Shen, Liu, & Shann, 2015), and Q-learning (Zhang et al., 2019) techniques in legal domain. However, there are also research concerning the pervasive use of AI by businesses poses new types of challenges to laws and social policy, such as consumer law and policy (Helveston, 2015). But our is more like the ones that applying machine learning in legal domain tasks. Specifically, what we have done is applying the improved and specific event extracting algorithm in the assisting judges handling divorce cases more efficiently and effectively.

**Event Extraction**: At present, the technologies of Information retrieval (IR), Information Extraction (IE), Text Mining are widely used in various domains (e.g., financial (Yang et al., 2018), news (Rasouli, Zarifzadeh, & Rafsanjani, 2020), personal life (Khodabakhsh, Kahani, & Bagheri, 2020) and education (Wang, Chen, & Lin, 2020)) to solve actual business problems, increase work efficiency, improve the competitiveness of enterprises and so on. Event Extraction is a subtask of Information Extraction (IE), which was initially driven by the Message Understanding Conferences (MUC) series and promoted by the Automated Content Extraction (ACE) and Biomedical Natural Language Processing Shared Tasks (BioNLP STs) greatly. More recently the Text Analysis Conferences (TAC) also propose EE task based on ACE.

There are many ways to extract events. With the discovery that the neural network technology can automatically learn the hidden data features, the Convolutional Neural Network (CNN) technology is used to determine the trigger chunks and determine which role each argument can match (Chen et al., 2015). Jagannatha & Yu (2016) extracts event instances from health records with Bidirectional Recurrent Neural Networks (Bi-RNNs). Further, Feng, Qin, & Liu, (2018) develops a hybrid neural network (a CNN and an RNN) to capture both sequence and chunk information from sentences, and use them to train an event detector that does not depend on any handcrafted features. Fei, Ren, & Ji, (2020) integrate RNN and CRFs as a unified framework, namely RecurCRFs, for biomedical event trigger detection. For the shortcoming of event extraction where it relies heavily on the quantity and quality of artificial identification data, and the training model is difficult to apply to new areas, Huang et al. (2018) put forward a kind of zero-shot learning method. Wiedmann (2017), Yang et al. (2018) and Zeng et al. (2018) all propose the Distance Supervision (DS) method which uses the entries in an existing structured database (such as FreeBase, or other relational database storage for specific scenarios) to mark the corresponding source natural language text to generate training samples. Liu, Luo & Huang (2018) and Nguyen & Grishman (2018) employ the graph convolutional network (GCN) to make use of graph structures of the origin sentences to detect event. Aiming at the shortcoming of ACE 2005 data collection, which is the uneven distribution of data of different event types and ambiguity of words in a single language, Liu, Chen, Liu & Zhao (2018) proposes a novel method that uses multilingual version of data to supplement each other, which enhance the characteristics of a small number of training samples and avoid ambiguity of single-language words.

## 6. Conclusion

Currently, judges need to manually find the key information and detect conflicts from textual materials delivered by both the plaintiff and the defendant while preparing the trial of divorce cases. For improving the efficiency of handling divorce cases, we propose an event-extracting-based approach to automatically extract core events and detect dispute issues. The proposed two-round-labeling strategy can solve the problem of sharing arguments or trigger words in multi events properly. Besides, we implement the Judicial Intelligent Assistant (JIA) system based on the approach. Experimental results show that the system can accurately obtain the key events of cases and help judges quickly identify the dispute issues. However, there are still a lot of work to be done in the future for make the system better. For example, the system is currently only applicable to divorce dispute. While applying to other type of civil cases, data for training the event extracting model should be manually annotated again, which is time-consuming and labor-intensive. Besides, not all labeled cases would eventually positively affect the model performance. So, as future works, we would like to explore how to automatically choose cases that would be useful in training the event extracting model from numerous cases, how to improve the event extraction performance, and how to make further use of the extracted events, such as constructing the behavior graph of parties.

**Acknowledgments**

This work was supported by the National Key R&D Program of China (2016YFC0800803), the National Natural Science Foundation of China (No. 61802167), Open Foundation of State key Laboratory of Networking and Switching Technology (Beijing University of Posts and Telecommunications) (SKLNST-2019-2-15), and the Fundamental Research Funds for the Central Universities.